\documentclass{article}

\usepackage[preprint,nonatbib]{neurips_2024}

\usepackage[utf8]{inputenc} % allow utf-8 input
\usepackage[T1]{fontenc}    % use 8-bit T1 fonts
\usepackage{hyperref}       % hyperlinks
\usepackage{url}            % simple URL typesetting
\usepackage{booktabs}       % professional-quality tables
\usepackage{amsfonts}       % blackboard math symbols
\usepackage{nicefrac}       % compact symbols for 1/2, etc.
\usepackage{microtype}      % microtypography
\usepackage{xcolor}         % colors

\usepackage{amsmath,amsfonts,amssymb}
\usepackage{algorithmic}
\usepackage{algorithm}
\usepackage{array}
\usepackage{caption}
\usepackage{subcaption}
\usepackage{textcomp}
\usepackage{stfloats}
\usepackage{verbatim}
\usepackage{graphicx}
\usepackage{comment}
\usepackage{color}
\usepackage{bbm}
\usepackage{multirow}
\usepackage{setspace}
\usepackage{wrapfig}

\title{Remembering Transformer For Continual Learning}

\newcommand*\samethanks[1][\value{footnote}]{\footnotemark[#1]}
\author{\normalsize Yuwei Sun\textsuperscript{1,2}, Ippei Fujisawa\textsuperscript{1}, Arthur Juliani\textsuperscript{3}, Jun Sakuma\textsuperscript{2,4,\thanks{Equal advising; Corresponding author: yuwei\_sun@araya.org }}, Ryota Kanai\textsuperscript{1,\samethanks}\\
\normalsize\textsuperscript{1}Araya, \textsuperscript{2}RIKEN AIP, \textsuperscript{3}Microsoft Research, \textsuperscript{4}Tokyo Institute of Technology\\
}

\begin{document}

\maketitle

\begin{abstract}
Neural networks encounter the challenge of Catastrophic Forgetting (CF) in continual learning, where new task learning interferes with previously learned knowledge. Existing data fine-tuning and regularization methods necessitate task identity information during inference and cannot eliminate interference among different tasks, while soft parameter sharing approaches encounter the problem of an increasing model parameter size. To tackle these challenges, we propose the Remembering Transformer, inspired by the brain's Complementary Learning Systems (CLS). Remembering Transformer employs a mixture-of-adapters architecture and a generative model-based novelty detection mechanism in a pretrained Transformer to alleviate CF. Remembering Transformer dynamically routes task data to the most relevant adapter with enhanced parameter efficiency based on knowledge distillation. We conducted extensive experiments, including ablation studies on the novelty detection mechanism and model capacity of the mixture-of-adapters, in a broad range of class-incremental split tasks and permutation tasks. Our approach demonstrated SOTA performance surpassing the second-best method by 15.90\% in the split tasks, reducing the memory footprint from 11.18M to 0.22M in the five splits CIFAR10 task.
\end{abstract}

\section{Introduction}

\begin{wrapfigure}{r}{0.46\linewidth}
\vspace{-12pt}
\centering
        \includegraphics[width=0.9\linewidth]{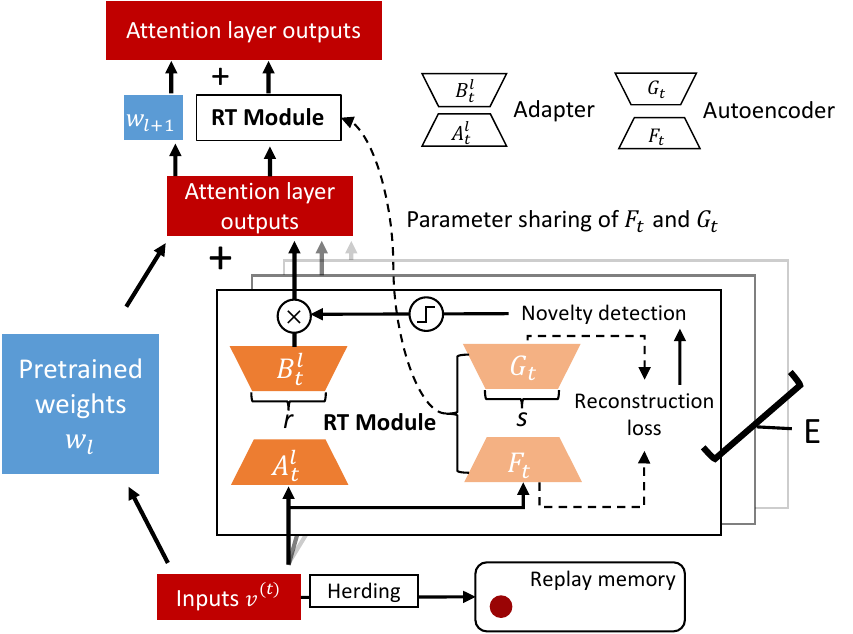}
        \caption{Remembering Transformer leverages the mixture-of-adapters that are sparsely activated with a novelty detection mechanism.}
        \vspace{-15pt}
        \label{fig:scheme1}
\end{wrapfigure}

The Catastrophic Forgetting (CF) problem arises in the sequential learning of neural networks, wherein the new task knowledge tends to interfere with old knowledge, with previously learned tasks forgotten. The ability to learn a new task without interfering with previously learned ones is of great importance to achieve continual learning. An intuitive approach to CF is data fine-tuning \cite{wangpers}. A model alleviates the forgetting of previous tasks by replaying and retraining on old task samples during learning a new task. Nevertheless, as the number of encountered tasks increases, it necessitates the storage of a large number of old task samples. In addition, the data fine-tuning based on memory replay usually cannot eliminate the CF problem leading to suboptimal performance.

Biological neural networks exhibit apparent advantages in continual learning via the Complementary Learning Systems (CLS) \cite{mcgaugh2000memory,kumaran2016learning}. In CLS, the Hippocampus rapidly encodes task data and then consolidates the task knowledge into the Cortex by forming new neural connections. The hippocampus has developed a novelty detection mechanism to facilitate consolidation by switching among neural modules in the Cortex for various tasks \cite{kafkas2018memory,gomez2022synaptic}. To this end, we propose a novel Remembering Transformer inspired by the CLS in the brain. Our approach consolidates a mixture-of-adapters architecture with generative routing to tackle the challenges in conventional continual learning methods (see Figure \ref{fig:scheme1}). Notably, the mixture-of-adapters architecture incorporates and jointly trains low-rank adapters with a pretrained Transformer. Then, these adapters are selectively activated for parameter-efficient fine-tuning through a novelty detection mechanism, where each task's samples are encoded in specific autoencoders for effective routing. During inference, the autoencoders, each encoding different task knowledge, predict routing weights for the various adapters, thus allocating an input sample to the most relevant adapter.

Moreover, we leverage low-rank autoencoders to facilitate novelty detection based on the magnitudes of reconstruction losses of various autoencoders, with the current task data as input. The reconstruction losses represent the similarity between the current task and the old task knowledge encoded in these autoencoders. %Remembering Transformer accommodates a new task leveraging a generative model for effective routing and an adapter for parameter-efficient fine-tuning. 
In a more challenging setting of limited model parameter capacity, we assume there is a limit for the number of adapters added to the Transformer. To tackle this challenge, we further propose the adapter fusion approach based on knowledge distillation \cite{distill} by distilling knowledge from the most relevant old adapter into a newly learned adapter. The knowledge distillation leverages a small set of old task replay memory and trains the new adapter on the probability distribution of the replay memory samples. Consequently, we demonstrate the superiority of Remembering Transformer in terms of enhanced task accuracy and parameter efficiency compared to a wide range of conventional continual learning methods. The results indicate that the proposed Remembering Transformer can achieve competitive performance even with the limited model capacity.

Overall, our main contributions are three-fold:

1) We propose the Remembering Transformer inspired by the Complementary Learning Systems to tackle the catastrophic forgetting problem in split tasks and permutation tasks. We investigate two challenging real-world scenarios in continual learning: class-incremental learning without task identity information and learning with limited model parameter capacity (Section \ref{sec:routing}). 

2) We propose the adapter fusion to distill old task knowledge from existing adapters and aggregate relevant adapters, enhancing parameter efficiency when learning with limited model parameter capacity (Section \ref{sec:fusion}). 

3) The empirical experiment results demonstrate the superiority of Remembering Transformer compared to conventional methods including Feature Translation and Representation Expansion, in terms of task accuracy and parameter efficiency in various continual learning tasks (Section \ref{sec:experiment}). 

The remainder of this paper is structured as follows. Section 2 reviews the most relevant work on continual learning. Section 3 demonstrates the essential definitions, assumptions, and technical underpinnings of the Remembering Transformer. Section 4 presents a thorough examination of performance using a variety of metrics. Section 5 concludes findings and gives out future directions.

\section{Related work}

This section provides a summary and comparison of relevant research on continual learning. To tackle the catastrophic forgetting (CF) problem, there are approaches including fine-tuning, generative replay, regularization, and soft parameter sharing. The fine-tuning method retrains the model with data from previous tasks when training on new task data \cite{wangpers}. Generative replay aims to train a generative model to reconstruct previous task data for the retraining \cite{huover,aljuexpe}. However, with the increasing number of tasks, it becomes less and less feasible to either store or learn a generative model for the retraining on the previous task data. A recent study \cite{huover} showed that the learned generative model was encountered with the CF problem as well, leading to the degraded quality of reconstructed task data. Additionally, fine-tuning on each observed task is computationally expensive for continual learning. To tackle these problems, regularization and soft parameter sharing methods selectively update model parameters without disturbing parameters important for old tasks \cite{kirkpatrick2017overcoming,ahn2019uncertainty}. The soft parameter sharing method leverages a modular architecture to learn task-specific neural module parameters \cite{jacoadap,ermicont,aljuexpe}. The isolation of parameters alleviates negative interference among various tasks. However, the soft parameter sharing method is inefficient in terms of model parameters and typically requires the task identity information for selecting relevant parameters. 

In contrast, the Remembering Transformer leverages a novelty detection method to accurately infer relevant neural modules without requiring any task identity information. It significantly reduces the model parameter size of each module based on a mixture-of-adapters architecture, wherein a set of task-specific linear transformation matrices is employed in each attention layer of a pre-trained Transformer model. By adaptively learning, activating, and fusing these neural modules, it significantly reduces the memory cost for continual learning.

\section{Methodology}
\label{sec:method}

\begin{figure}[t]
        \centering
        \includegraphics[width=0.9\linewidth]{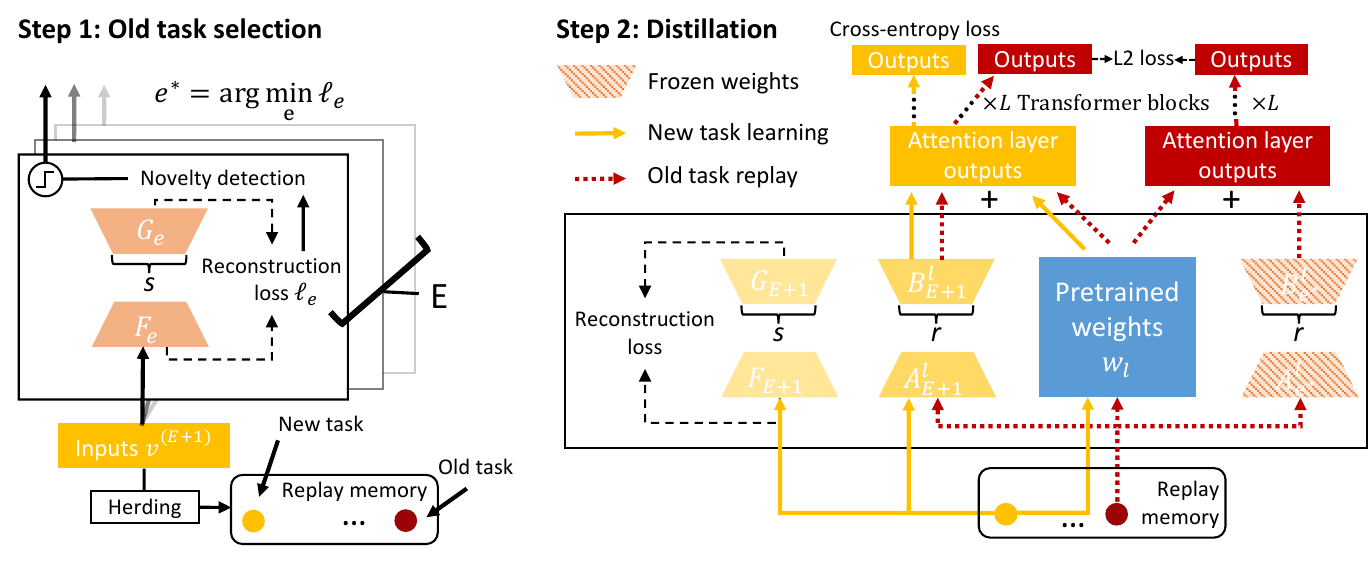}
        \caption{Adapter fusion based on knowledge distillation with a limited capacity $E$. We update the $E+1$-th adapter $\{B^l_{E+1}, A^l_{E+1}\}_{l=1}^L$ using new task data and the soft probability output of the old task replay.}
    \label{fig:scheme2}
\end{figure}

In this section, we delve into a comprehensive exploration of the assumptions and the proposed Remembering Transformer’s technical underpinnings. These include the incorporation of mixture-of-adapters in Transformer models, a generative model-based novelty detection for expert routing, and adapter fusion based on knowledge distillation (see Figure \ref{fig:scheme2}). The adapter fusion enables efficient knowledge retrieval from relevant previously learned adapters while learning new tasks, preventing catastrophic forgetting of previous tasks.

\subsection{Continual learning task setting}

We aim to tackle continual learning on a sequence of tasks $\{T^{(1)}, T^{(2)}, \dots, T^{(N)}\}$, where $T^{(t)} = \{x_i^t, y_i^t\}_{i=1}^{K_t} \subset T$ is a labeled dataset including $K_t$ pairs of instances $x^{t}$ and their corresponding labels $y^{t}$. We investigate the most challenging class-incremental learning setting in continual learning, where each task \(T^{(t)}\) is defined by a non-overlapping subset of a dataset \(T^{(t)}\subset \{x_i, y_i\}_{i=1}^{K}\) where $K=\sum_{t=1}^N K_t$. \(Y^t \subset Y\) is changing over time, i.e., \(Y^{i} \cap Y^{j}=\emptyset\,\) \(\forall i \neq j\). The posterior $P(X|Y)$ is consistent across tasks $T^{(t)}\subset T$. During inference, we evaluate on $P(X) \sim \frac{1}{K}\sum_{t=1}^N P(X|Y) P_t(Y)$ without the task identity information $t$.

\subsection{Vision Transformers with mixture-of-adapters}

Our method is inspired by the Mixture-of-Experts (MoE) \cite{scale, vit, sparse}, with diverse neural modules constituting a comprehensive neural network. Different parts of the neural network learn and adapt independently without disrupting the knowledge that other parts have acquired, thus mitigating interference between tasks. The intuition is to learn a diverse collection of neural modules based on a pretrained Vision Transformer (ViT) and adaptively select the most relevant neural module for adapting to various continual learning tasks.  

ViT partitions an image into a sequence of non-overlapping patches. Let $x\in\mathbb{R}^{H\times W\times C}$ be an image input, where $(H,W)$ is the resolution of the image and $C$ is the number of channels. $x$ is separated into a sequence of 2D patches $x_p\in \mathbb{R}^{N\times (P^2\cdot C)}$, where $(P,P)$ is the resolution of each image patch and $\frac{HW}{P^2}$ is the number of patches. These patches are mapped to tokens $v_p\in\mathbb{R}^{\frac{HW}{P^2}\times D}$ with a learnable linear projection. A learnable 1D position embedding $v_0\in\mathbb{R}^{D}$ is prepended to the tokens to retain positional information, resulting in $v\in\mathbb{R}^{(\frac{HW}{P^2}+1)\times D}$.

Moreover, we employ the soft parameter sharing method to enhance ViT's ability for continual learning, alleviating catastrophic forgetting. Notably, for each task \( T^{(t)} \subseteq T \), we utilize the Low-Rank Adaptation (LoRA) method \cite{hu2021lora} for efficient fine-tuning using low-rank decomposition matrices. We employ the trainable decomposition matrices to the attention weights of ViT's different layers. For each attention layer $l$, linear transformations are applied to the query, key, value, and output weights ($W_l^Q, W_l^K, W_l^V, W_l^O$). For $W_l \in \mathbb{R}^{D\times D}$ $\in$ ($W_l^Q, W_l^K, W_l^V, W_l^O$), the parameter update $\Delta W_l$ is then computed by $W_l + \Delta W_l = W_l + B^lA^l$, where $A^l\in \mathbb{R}^{r\times D}, B^l \in \mathbb{R}^{D\times r}$, and $r$ is a low rank $r \ll D$. For an input token $v^{t}$, the output of the $l$th attention layer is $\hat{W_l}v^{t} + \Delta W_lv^{t} = \hat{W_l}v^{t} + B^lA^lv^{t}$ where $\hat{\cdot}$ indicates untrainable parameters. For each task $T^{(t)}$, $\theta^t_{\text{adapter}} = \{A_t^l, B_t^l\}_{l=1}^L$ is added to the pretrained ViT model $\{\hat{\theta}_{\text{ViT}},\theta^t_{\text{adapter}}\}$ where $L$ represents the total number of attention layers. We formulate the optimization of the adapter based on task $t$'s dataset $\{v_i^t,y_i^t\}_{i=1}^{K_t}$ as follows: $\hat{\theta}_{\text{adapter}}^{t}\leftarrow\underset{\theta_{\text{adapter}}^{t}}{\text{arg min }}-\sum_{i=1}^{K_{t}} f(v_i^{t};\{\hat{\theta}_{\text{ViT}},\theta_{\text{adapter}}^{t}\})\text{log}(y_i^t).$ Then, $\hat{\theta}_{\text{adapter}}^{t}$ is added to the list of adapters $\{{\theta}_{\text{adapter}}^{1},{\theta}_{\text{adapter}}^{2}, ..., {\theta}_{\text{adapter}}^{t-1}\}$ to form the mixture-of-adapters architecture.

\subsection{Generative model-based novelty detection for expert routing}
\label{sec:routing}

Utilizing the input tokens, our goal is to adaptively activate the most relevant adapters within the mixture-of-adapters framework. We propose a novelty detection mechanism based on generative models for effective adapter routing. In particular, a routing neural network outputs gating weights $W_g \in \mathbb{R}^E$ that indicate the probability of employing a specific adapter for a given task, where $E$ represents the number of existing adapters. Conventional routing functions in Mixture-of-Experts (MoE) usually struggle with the more complex continual learning tasks, where sequential task inputs necessitate a continual update of the gating weights leading to forgetting of the routing neural networks. 

A generative model encodes tokens from a specific task and assesses the familiarity of a new task in relation to the encoded knowledge of the old task. Notably, to facilitate efficient routing, we employ a low-rank autoencoder $\theta_{\text{AE}}$ that consists of an encoder and decoder, each of which leverages one linear transformation layer $F \in \mathbb{R}^{s\times D}$ and $G \in \mathbb{R}^{D\times s}$ where $s$ is a low rank $s \ll D$. For each task $t$, we train the AE to encode the embedding layer output $V^{t}$, which are flattened and passed through a Sigmoid activation $\sigma(\cdot)$, $\hat{V^{t}} = G_tF_t\sigma(V^{t})$. The Sigmoid activation constraints the reconstructed tokens from the AE between the range of 0 and 1 to facilitate learning. Then, the AE is updated based on the mean squared error loss in Eq. (\ref{eq:ae}). 
%Additionally, for enhanced parameter efficiency, a routing model $\theta_{\text{AE}}$ is globally shared across different attention layers.
\begin{equation}
\label{eq:ae}
\begin{aligned}
\ell_t(\theta_{\text{AE}}^t) = ( \sigma(V^{t}) - \hat{V^{t}})^2,\,
\hat{\theta}_{\text{AE}} =\underset{\theta_{\text{AE}}^t}{\text{arg min }}\ell_t.\\
\end{aligned}
\end{equation}
During inference, the collection of AEs $\hat{\theta}_{\text{AE}}^e \in \{\hat{G}_e, \hat{F}_e\}_{e=1}^E$ trained on various tasks provide estimates of the input tokens $v$'s novelty in relation to the previously learned tasks based on the computed reconstruction loss $\ell_e(v)$. In particular, when the input closely resembles an old task learned by an existing AE, the reconstruction loss for the input task by this AE tends to be low. Consequently, the old task $e^*$ corresponding to the AE with the minimum reconstruction loss is the most relevant to the input task. Then, the adapter $\theta^{e^*}_{\text{adapter}}$ is leveraged to adapt the pretrained ViT for tackling the input task. We devise the proposed novelty detection mechanism as follows: 
\begin{equation}
\label{eq:gate}
\begin{aligned}
\ell_e = ( \sigma(v) - \hat{G}_e\hat{F}_e\sigma(v))^2,\,
e^*=\underset{e}{\text{arg min }}\ell_e,\\
W_g(e) = \begin{cases} 
      1 & \text{if } e = e^* \\
      0 & \text{otherwise}.
      \end{cases}
\end{aligned}
\end{equation}
The output $v_{l}$ of the $l$th attention layer is then computed by $\hat{W_l}v_{l-1} + \sum_{e=1}^{E} W_g(e)B^l_eA^l_ev_{l-1}$, where $v_{l-1}$ is the output of the previous $l-1$th attention layer. Additionally, for enhanced parameter efficiency, the routing model $\{\theta_{\text{AE}}^t\}_{t=1}^{N}$ is globally shared across different attention layers.

\subsection{Adapter fusion based on knowledge distillation}
\label{sec:fusion}

Remembering Transformer leverages the mixture-of-adapters architecture and the low-rank generative routing for enhanced parameter efficiency in continual learning. However, with the increasing number of tasks, it is still encountered with increasing parameters. To further enhance its parameter efficiency, we explore a scenario where the number of adapters is constrained. We then propose a novel adapter fusion approach within the mixture-of-adapters framework to identify and aggregate resembling adapters based on knowledge distillation \cite{distill}. This involves transferring the knowledge of a selected old adapter $\theta_{\text{adapter}}^{e^*}$ to the new adapter $\theta_{\text{adapter}}^{t}$ by replaying old task samples when the model capacity $E$ is reached, i.e., when $t\leq E+1$ (see Figure \ref{fig:scheme2}). 

Furthermore, to reduce the computational cost of knowledge distillation, we leverage the Herding method \cite{wellherd09,rebuicar17} to obtain a small set of the most representative replay samples of each task during training. In particular, given tokens from a specific class $y \in Y^t$ of task $t$, i.e., $V^{t}_y = \{v^{t}_{y,1}, v^{t}_{y,2}, \dots, v^{t}_{y,K_t^y}\}$ where $K_t^y$ is the total number of samples from class $y$, we compute the class center $\mu_y$ based on the latent representations learned by the autoencoder $\hat{\theta}_{\text{AE}}^t:=\{\hat{G}_t, \hat{F}_t\}$. The class center in the latent space is $\mu_y\leftarrow\frac{1}{K_t^y}\sum_{i=1}^{K_t^y}\hat{F}_t\sigma(v^{t}_{y,i})$. Then, the distance of each token's latent representation to the class center $||\mu_y-\hat{F}_t\sigma(v^{t}_{y,i})||$ is ranked in the ascending order. For each class $y \in Y^t$, we select and add the top-$\frac{M}{U}$ tokens ($U$ is the number of classes and $M$ is the task replay memory size) with the least distance to the class center in the latent space to the replay memory $\Xi_t$ of task $t$, i.e., $\Xi_t \leftarrow \{\text{top-}\frac{M}{U}(v^{t}_{y,i})\}_{y\in Y^t}$.

\begin{algorithm}[t]
%\setstretch{1.2}
\caption{Remembering Transformer with Adapter Fusion}
\label{algo:lora}
\begin{algorithmic}[1]
    \FOR{each task $t = 1, 2,\dots,N$}
            \STATE Initialize a new autoencoder $\theta_{\text{AE}}^{t}:\{G_t,F_t\}$.
            \STATE $\ell_{t} = \frac{1}{K_t}\sum_{i=1}^{K_t} (\sigma(v^{t}_{i}) - GF\sigma(v^{t}_{i}))^2.$
            \STATE $\hat{\theta}_{\text{AE}}^{t}=\underset{\theta_{\text{AE}}^{t}}{\text{arg min }}\ell_{t}.$
            \STATE Add $\hat{\theta}_{\text{AE}}^{t}$ to the list of AE $\{{\theta}_{\text{AE}}^{1},{\theta}_{\text{AE}}^{2},...,{\theta}_{\text{AE}}^{t-1}\}$.
            \STATE For $y \in Y^t$,  $\mu_y \leftarrow\frac{1}{K_t^y}\sum_{i=1}^{K_t^y}\hat{F}_t\sigma(v^{t}_{y,i})$.
            \STATE $\Xi_t \leftarrow \{\text{top-}\frac{M}{U}(v^{t}_{y,i})\}_{y\in Y^t}$ by ranking $||\mu_y - \hat{F}_t\sigma(v^{t}_{y,i})||$ in the ascending order.
            \STATE Add $\Xi_t$ to the replay memory $\{\Xi_1,\Xi_2,...,\Xi_{t-1}\}$.  
            \FOR{each AE $e=1,2,...,t-1$ in the AE list}
                   \STATE $\ell_e = (\sigma(v_{i}^t) - \hat{G_e}\hat{F_e}\sigma(v_{i}^t))^2.$\\
            \ENDFOR
            \STATE Obtain the index of the most relevant AE: $e^*=\underset{e}{\text{arg min }}\ell_e$.
            \IF{$t>E$ }
            \STATE Replay old task $e^*$ samples from the memory: $\{v^{e^*}_i\}_{i=1}^{M} \leftarrow \Xi_{e^*}$.
            \STATE $\ell_{\text{L2}}(\theta_{\text{adapter}}^t)=\sum_{i=1}^{M}\left(f(v^{e^*}_i;\{\hat{\theta}_{\text{ViT}},\hat{\theta}_{\text{adapter}}^{e^*}\}) - f(v^{e^*}_i;\{\hat{\theta}_{\text{ViT}},\theta_{\text{adapter}}^{t}\})\right)^2$.
            \STATE $\ell_{\text{CE}}(\theta_{\text{adapter}}^t)=-\sum_{i=1}^{K_{t}} f(v_i^{t};\{\hat{\theta}_{\text{ViT}},\theta_{\text{adapter}}^{t}\}) \log(y^{t}_i)$.
            \STATE $\hat{\theta}_{\text{adapter}}^{t}\leftarrow\underset{\theta_{\text{adapter}}^{t}}{\text{arg min }}\alpha\cdot\ell_{\text{CE}} + (1-\alpha)\cdot\ell_{\text{L2}}.$
            \STATE The old adapter is removed and the samples of task $e^*$ are distributed to the newly learned adapter $t$: Gate$(e^*) \leftarrow t$.
            \ELSE
            \STATE Initialize a new adapter $\theta_{\text{adapter}}^{t} = \{A_{t}^l, B_{t}^l\}_{l=1}^L$.
            \STATE $\hat{\theta}_{\text{adapter}}^{t}\leftarrow\underset{\theta_{\text{adapter}}^{t}}{\text{arg min }}-\sum_{i=1}^{K_{t}} f(v_i^{t};\{\hat{\theta}_{\text{ViT}},\theta_{\text{adapter}}^{t}\})\text{log}(y_i^t).$
            \STATE Add $\hat{\theta}_{\text{adapter}}^{t}$ to the list of adapters $\{{\theta}_{\text{adapter}}^{1},{\theta}_{\text{adapter}}^{2}, ..., {\theta}_{\text{adapter}}^{t-1}\}$.
            \ENDIF
    \ENDFOR
\end{algorithmic}
\end{algorithm}

The old adapter \( e^* \) for knowledge distillation is selected based on the novelty detection mechanism in Eq. (\ref{eq:gate}). We determine the most relevant old adapter for the new task by identifying the minimum reconstruction loss. We compute the soft probability output $f(\Xi_{e^*};\{\hat{\theta}_{\text{ViT}},\hat{\theta}_{\text{adapter}}^{e^*}\})$ of the old adapter by replaying samples $\Xi_{e^*}$ from the memory. Then, the new adapter $\theta_{\text{adapter}}^{E+1}$ is trained using its task data $\{V^{E+1},Y^{E+1}\}$ and the soft probability output of the old task $\{\Xi_{e^*}, f(\Xi_{e^*};\{\hat{\theta}_{\text{ViT}},\hat{\theta}_{\text{adapter}}^{e^*}\})\}$. Notably, we employ the cross-entropy loss $\ell_\text{CE}(\cdot)$ and the L2 loss $\ell_\text{L2}(\cdot)$ for training on the new and old tasks, respectively. We devise the optimization process for the adapter fusion as follows: 
\begin{equation}
\begin{aligned}
    \ell_\text{CE}(\theta_{\text{adapter}}^{E+1}) = -\sum_{i=1}^{K_{E+1}} f(v_i^{E+1};\{\hat{\theta}_{\text{ViT}},\theta_{\text{adapter}}^{E+1}\}) \log(y^{E+1}_i),\\
    \ell_\text{L2}(\theta_{\text{adapter}}^{E+1}) = \sum_{i=1}^{M}\left(f(v^{e^*}_i;\{\hat{\theta}_{\text{ViT}},\hat{\theta}_{\text{adapter}}^{e^*}\}) - f(v^{e^*}_i;\{\hat{\theta}_{\text{ViT}},\theta_{\text{adapter}}^{E+1}\})\right)^2,\\
    \hat{\theta}_{\text{adapter}}^{E+1} = \underset{\theta_{\text{adapter}}^{E+1}}{\text{arg min }}\alpha\cdot\ell_\text{CE}+ (1-\alpha)\cdot\ell_\text{L2},
\end{aligned}
\end{equation}
where $\alpha$ is a coefficient to balance the two loss items. 

Additionally, after adapter fusion, the old adapter is removed, and the samples of task $e^*$ are distributed to the newly learned adapter $E+1$ for inference, i.e., Gate$(e^*) \leftarrow E+1$ where Gate($\cdot$) is a function to project the result of novelty detection to the defined new gate. Note that Gate($\cdot$) is initialized with an identity projection $\text{Gate}_{\text{init}}(e) = e$. We can formulate the routing weights as follows: $W_g(e) = \begin{cases} 
      1 & \text{if } e = \text{Gate}(e^*) \\
      0 & \text{otherwise}.
      \end{cases}$

We devise the algorithm for the Remembering Transformer in Algorithm \ref{algo:lora}.

\section{Experiments}
\label{sec:experiment}

In this section, we provide a detailed description of the datasets, model architectures, and metrics used in the experiments. An extensive empirical evaluation including an ablation study is performed based on a wide range of split and permutation tasks in continual learning. Moreover, we investigate the Remembering Transformer's performance when its capacity of adapters is limited. The results demonstrate that the Remembering Transformer achieves SOTA performance while retaining great parameter efficiency compared with conventional methods.

\subsection{Settings}
\paragraph{Datasets}

We evaluated the model performance based on continual learning datasets adapted from CIFAR10, CIFAR100 \cite{cifar10}, and Permuted MNIST \cite{van2020brain} tasks. We investigated two types of continual learning datasets: (1) the split task where each dataset is split into several subsets of equal numbers of classes, with each subset as a task. For example, we divide the CIFAR-10 dataset into five tasks. The first task consists of digits (classes) 0 - 1 and the second task consists of digits (classes) 2 - 3 and so on. We denote the task of CIFAR-10 in five splits as CIFAR10/5. (2) the permutation task where the input pixels of an image in the training and test data are shuffled with a random permutation, with a different permutation for each task. 

\paragraph{Metrics}

For the split tasks, we measure the average task accuracy over all tasks after training, i.e., $\text{Acc}= \frac{1}{N}\sum_{i=1}^N \text{Acc}_i$ where \( \text{Acc}_i \) is the accuracy on the \( i \)th task after learning the final task $T^{(N)}$. Moreover, we evaluate the memory footprint of a method based on its trainable model parameter size. For the permutation tasks, we evaluate the average task accuracy each time the model learns a new permutation. We then report the mean and standard deviation of three individual experiments with random seeds, with each seed resampling the splits to avoid any favorable class split and resampling the permutation patterns to avoid any favorable permutation.

\paragraph{Hyperparameters}

We employed the base-size Vision Transformer pretrained on ImageNet-21k, with a resolution of 224x224 pixels and a patch size of $16\times16$. For the detailed architecture, we followed the default author-suggested settings \cite{vit}. We employed a two-layer autoencoder with a latent dimension of one for all split tasks (5-20 splits) and a four-layer autoencoder with a hidden dimension of 32 and a latent dimension of one for the permutation task (50 permutations). Moreover, the hyperparameters were chosen based on a grid search. Batch sizes of 128 and 1 were employed for training and test, respectively. We utilized the AdamW optimizer with $\beta_1=0.9$, $\beta_2=0.999$, and a weight decay of 0.01. We employed learning rates of 0.001 and 0.005 for updating the adapters and autoencoders, respectively. We employed a coefficient of 0.5 in the loss function of the knowledge distillation. We trained the model for 50 epochs in the CIFAR10 split task, 200 epochs in the CIFAR100 split tasks, 2000 epochs in the permuted-MNIST task. Additionally, we employed the average along the latent dimension of each token of the embedding layer output to train the autoencoders for enhanced parameter-efficiency. The autoencoders were trained once for 10 epochs at the beginning of each task. All experiments were based on PyTorch and four A100 GPUs. The code would be made publicly available.

\subsection{Class-incremental split tasks}

\paragraph{Average task accuracy}

We train all tasks for $\frac{\text{total epochs}}{\text{\# of tasks}}$ epochs before the model training on a different task. We conducted a comprehensive evaluation in the CIFAR10/5, CIFAR100/10, and CIFAR100/20 split tasks based on the average task accuracy, comparing to a wide range of conventional methods. Table \ref{tab:acc} demonstrated that Remembering Transformer significantly enhanced model performance in the challenging class-incremental learning without the task identity information. Remembering Transformer surpassed the second-best method, FeTrIL \cite{petit23}, by 15.90\%, establishing a SOTA performance.

\begin{table}[t]
\centering
\small
\caption{Average task accuracy in the split tasks.}
\begin{tabular}{lcccc}
\toprule
Method & \multicolumn{1}{c}{CIFAR10/5 (\%)} & \multicolumn{1}{c}{CIFAR100/10 (\%)} & \multicolumn{1}{c}{CIFAR100/20 (\%)} & \multirow{1}{*}{Average (\%)} \\ \hline
SupSup \cite{wortsman20} & 26.2 $\pm$ 0.46 & 33.1 $\pm$ 0.47 & 12.3 $\pm$ 0.30 & 23.87 \\
BI-R+SI \cite{van2020brain} & 41.7 $\pm$ 0.25 & 22.7 $\pm$ 0.81 & 19.1 $\pm$ 0.04 & 27.83 \\
HyperNet \cite{oswald20} & 47.4 $\pm$ 5.78 & 29.7 $\pm$ 2.19 & 19.4 $\pm$ 1.44  & 32.17 \\ 
MUC \cite{liu20} & 53.6 $\pm$ 0.95 & 30.0 $\pm$ 1.37 & 14.4 $\pm$ 0.93 & 32.67 \\
OWM \cite{zeng19} & 51.7 $\pm$ 0.06 & 29.0 $\pm$ 0.72 & 24.2 $\pm$ 0.11  & 35.63 \\
PASS \cite{zhu21} & 47.3 $\pm$ 0.97 & 36.8 $\pm$ 1.64 & 25.3 $\pm$ 0.81 & 36.47 \\
LwF \cite{liucvpr} & 54.7 $\pm$ 1.18 & 45.3 $\pm$ 0.75 & 44.3 $\pm$ 0.46 & 48.10 \\
iCaRL \cite{rebuffi17} & 63.4 $\pm$ 1.11 & 51.4 $\pm$ 0.99 & 47.8 $\pm$ 0.48 & 54.20 \\
Mnemonicsy \cite{liucvpr} & 64.1 $\pm$ 1.47 & 51.0 $\pm$ 0.34 & 47.6 $\pm$ 0.74 & 54.23 \\
DER++ \cite{buzzega20} & 66.0 $\pm$ 1.27 & 55.3 $\pm$ 0.10 & 46.6 $\pm$ 1.44 & 55.97 \\
IL2A\cite{zhu21neurips} & 92.0 $\pm$ 0.23 & 60.3 $\pm$ 0.14 & 57.9 $\pm$ 0.31 & 70.07 \\
CLOM \cite{kim22} & 88.0 $\pm$ 0.48 & 65.2 $\pm$ 0.71 & 58.0 $\pm$ 0.45 & 70.40 \\
SSRE \cite{zhu22cvpr} &  90.5 $\pm$ 0.61 & 65.0 $\pm$ 0.27 & 61.7 $\pm$ 0.15  & 72.40 \\
FeTrIL \cite{petit23}& 90.9 $\pm$ 0.38 & 65.2 $\pm$ 0.16 & 61.5 $\pm$ 0.73 & 72.53 \\
Ours & \textbf{99.3 $\pm$ 0.24} & \textbf{86.1 $\pm$ 2.09} & \textbf{79.9 $\pm$ 3.56} & \textbf{88.43} \\ 
\bottomrule
\end{tabular}
\label{tab:acc}
\end{table}

\paragraph{Accuracy curves}

\begin{figure}[t]
    \centering
    \includegraphics[width=\linewidth]{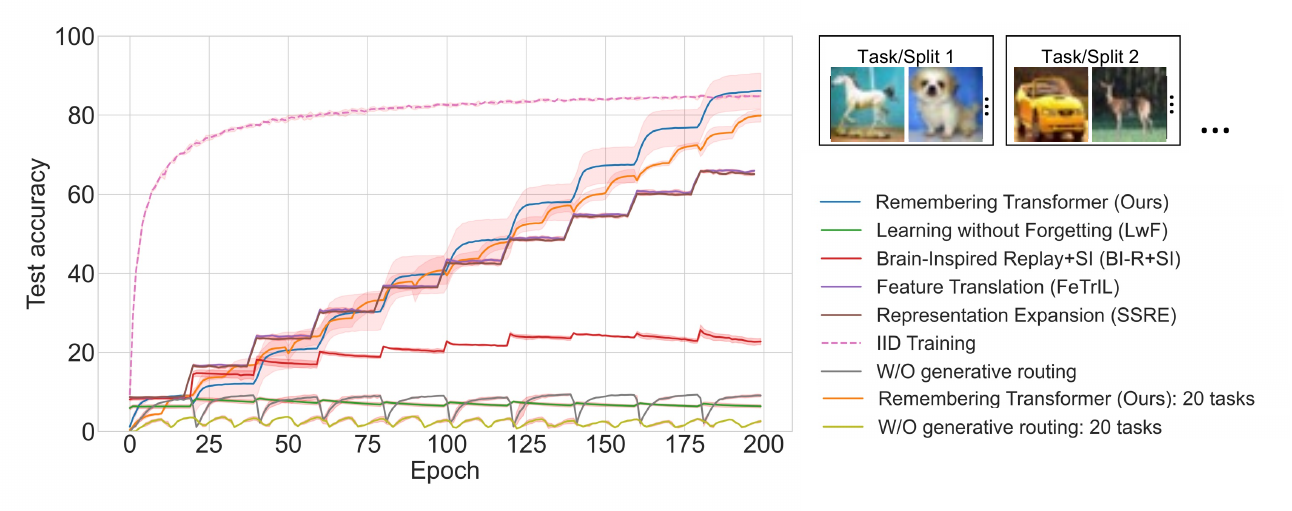}
    \caption{Test accuracy curves in the CIFAR100 tasks.}
    \vspace{-10pt}
    \label{fig:split}
\end{figure}

Figure \ref{fig:split} demonstrated the accuracy curves of the CIFAR100/10 and CIFAR100/20 split tasks. For comparison, we visualized the performance of FeTrIL \cite{petit23} and SSRE \cite{zhu22cvpr}. We added an oracle model based on an Independent and Identically Distributed (IID) training setting where tasks are defined as identical instances of the same dataset with all classes. This is equivalent to finetuning the pretrained model on the entire dataset based on a single adapter. The Remembering Transformer demonstrated greatly enhanced continual learning ability. In contrast, the existing methods encountered the forgetting problem when the task changed. The Remembering Transformer even outperforms the oracle model in the CIFAR100/10 task leveraging the proposed modular architecture for reduced interference among classes.

\paragraph{Memory footprint}

\begin{figure}[t]
    \centering
    \includegraphics[width=0.85\linewidth]{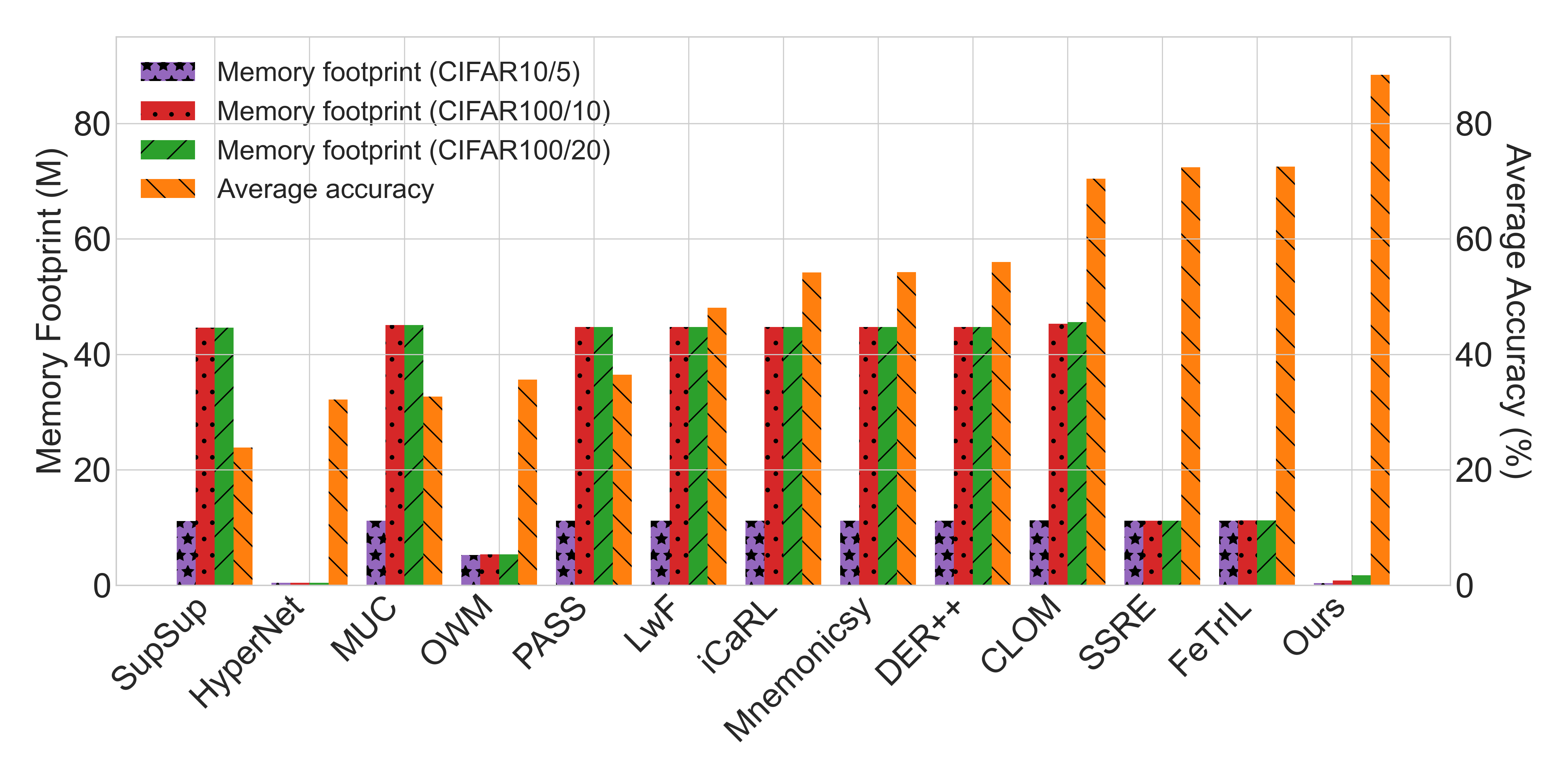}\vspace{-10pt}
    \caption{Memory footprint of the comparison models.}
    \vspace{-10pt}
    \label{fig:memory}
\end{figure}

We evaluated the memory footprint by measuring the trainable model parameter size. Figure \ref{fig:memory} demonstrated that the Remembering Transformer is parameter-efficient for continual learning compared to conventional methods, reducing the memory footprint in the CIFAR10 split task from 11.18M (FeTrIL \cite{petit23}) to 0.22M. Moreover, although OWM and HyperNet utilized smaller model sizes, they achieved much worse model performance. The Remembering Transformer achieved SOTA performance while retaining a small memory footprint.

\subsection{Ablation studies}

We studied the efficacy of the Remembering Transformer by conducting ablations on the generative routing mechanism and varying the model capacity of adapters.

\paragraph{Ablating the generative routing}

Remembering Transformer leverages the generative routing to facilitate sparse adapter activation. We investigated whether the Remembering Transformer with the generative routing ablated could tackle the CIFAR100 split tasks (see Figure \ref{fig:split}). %Moreover, we experimented on replacing the generative routing with a cross attention-based routing \cite{jacoadap,ermicont,disc}. 
The empirical results demonstrated that the mixture-of-adapters architecture alone could not adapt to the continual learning task. It highlighted the importance of the generative routing mechanism for a model continually learning new tasks without forgetting old knowledge. 
%Additionally, the cross attention-based routing was found to be prone to forgetting the adapter activation for old tasks.

\paragraph{Varying the model capacity}

The adapter fusion method leverages knowledge distillation to transfer old task knowledge to a newly learned adapter for reducing the memory footprint. To assess the model efficacy under a constrained adapter capacity, we varied the capacity by reducing the adapters to $E = {3,2}$ in the CIFAR10/5 split task while employing a replay memory size of $M = 512$. Table \ref{tab:model_capacity_accuracy} demonstrated that the Remembering Transformer could learn a number of tasks that were more than the adapters, retaining competitive performance. Moreover, we varied the replay memory size to $M = \{128, 256, 512\}$ when employing a model capacity of $E = 3$ (see Table \ref{tab:ablation}). Consequently, compared to FeTrIL, which obtained a performance of 90.9\% with a memory footprint of 11.18M (see Table \ref{tab:acc}), the Remembering Transformer achieved a performance of 93.2\% with a much smaller memory footprint of 0.22M, when $E = 3$ and $M = 512$.

\begin{table}[t]
\centering
\small
\begin{minipage}{0.475\linewidth}
\centering
\caption{Model capacity vs. test accuracy.}
\begin{tabular}{cc}
\toprule
\#Adapters (memory footprint) & Test accuracy (\%) \\
\midrule
5 (0.37M) & \textbf{99.3 $\pm$ 0.24} \\
3 (0.22M) & 93.2 $\pm$ 0.72 \\
2 (0.15M) & 87.1 $\pm$ 0.85 \\
\bottomrule
\end{tabular}
\label{tab:model_capacity_accuracy}
\end{minipage}
\hfill
\begin{minipage}{0.475\linewidth}
\centering
\caption{Replay memory size vs. test accuracy.}
\begin{tabular}{cc}
\toprule
\#Replay samples & Test accuracy (\%) \\
\midrule
512 & \textbf{93.2 $\pm$ 0.72}\\
256 & 90.5 $\pm$ 0.41\\
128 & 86.6 $\pm$ 0.26\\
\bottomrule
\end{tabular}
\label{tab:ablation}
\end{minipage}
\end{table}

\subsection{Permutation tasks}

\begin{figure}[t]
    \centering
    \includegraphics[width=0.88\linewidth]{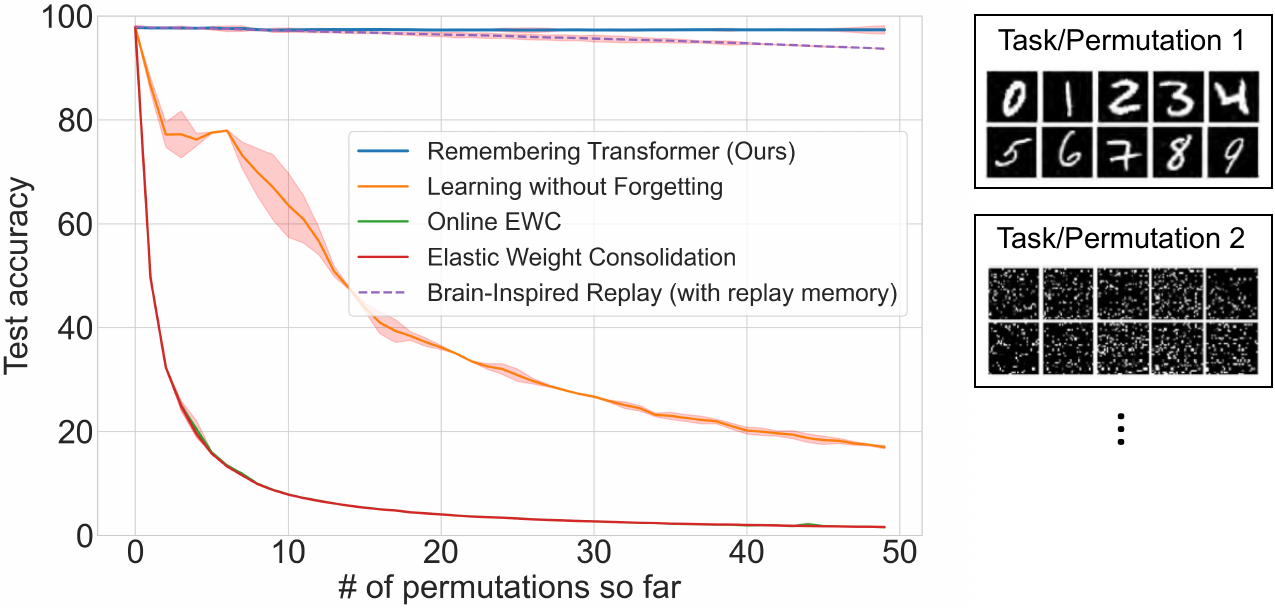}
    \caption{Test accuracy curves of the Remembering Transformer compared to the conventional methods for the permuted-MNIST task.}
    \label{fig:permute}
\end{figure}

\begin{wrapfigure}{r}{0.35\linewidth}
    \centering
    \vspace{-10pt}
    \includegraphics[width=0.82\linewidth]{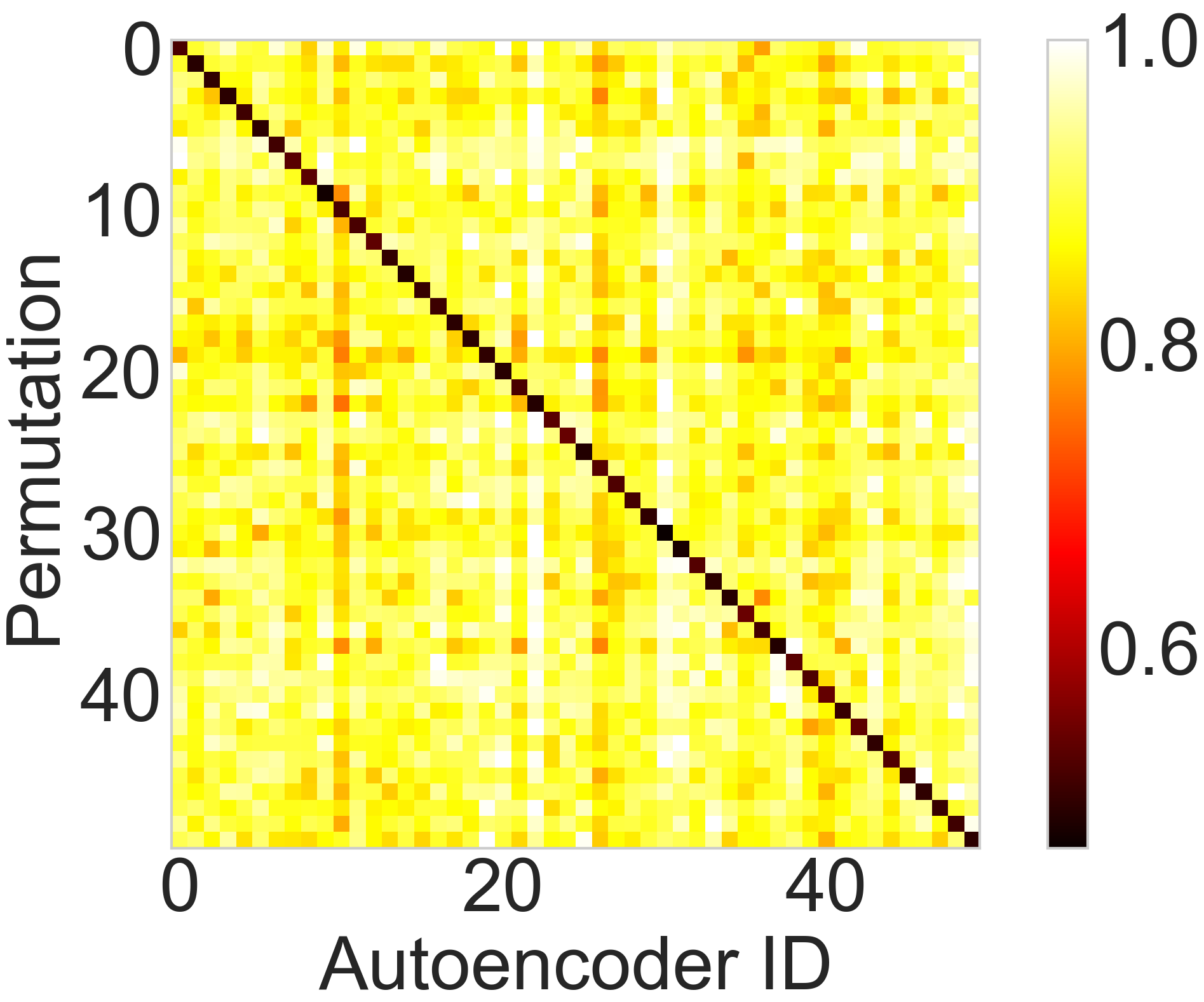}
    \caption{Heatmap of the reconstruction losses of autoencoders with respect to different tasks during inference time. %Compared to the permutation task in (b), the split task in (a) shows outputs of reconstruction losses that resemble each other more closely, thus making it more difficult to distinguish among various tasks. 
    %As we increase the autoencoder size to the same as the permutation task, the heatmap in (c) shows more distinct values, but with a small trade-off in its memory footprint.
    }
    \label{fig:heatmap}
\end{wrapfigure}

We evaluated the Remembering Transformer in the permuted-MNIST dataset. Each task consists of all 10 digits, with a specific permutation shift employed to the image pixels. During inference, the model is given samples featuring the same set of permutation patterns but lacks identity information on the various permutations. Notably, we employed a total of 50 different random permutation patterns and trained the model for 2000 epochs, with 40 epochs for each permutation. We evaluated the average task accuracy each time the model learned a new permutation. We demonstrated the Remembering Transformer's performance, comparing to a broad range of conventional methods for this dataset, including Learning without Forgetting \cite{li2017learning}, Elastic Weight Consolidation (EWC) \cite{kirkpatrick2017overcoming}, Online EWC \cite{schwarz2018progress}, and Brain-Inspired Replay \cite{van2020brain}. Figure \ref{fig:permute} demonstrated that the Remembering Transformer retained its performance while learning new permutations. While Brain-Inspired Replay also demonstrated competitive performance, it relies on model retraining using replay memory samples from all old tasks. In contrast, our method utilizes only the most relevant old task's replay samples during the adapter fusion when there is a constraint on the model capacity. Furthermore, we visualized the reconstruction losses between pairwise tasks and autoencoders during inference. A small magnitude of the reconstruction loss indicates a high degree of relevance of a task to the knowledge encoded in the specific autoencoder. For visualization, the raw losses were normalized within a range of 0 and 1. Figure \ref{fig:heatmap} demonstrated that the proposed generative routing method facilitated the adaptive selection of the most relevant adapter (along the diagonal) for tackling a specific task. 
%All tasks in the permuted MNIST and the vast majority of tasks in CIFAR100/20 were accurately correlated to the most relevant autoencoder. 
%As we increase the size of the autoencoder for CIFAR100/20 to match that of the permutation task, all tasks were successfully identified with the correct autoencoders, with a small trade-off in memory footprint.

\section{Limitations and conclusions}

The Remembering Transformer leverages the mixture-of-adapters architecture that is seamlessly integrated into conventional vision Transformers for parameter-efficient continual learning. The novelty detection method identifies task similarities and routes samples to the most relevant adapters. We demonstrated the superiority of the Remembering Transformer in various class-incremental split tasks and permutation tasks, resulting in SOTA performance with a small memory footprint. %Moreover, we showed that Remembering Transformer could achieve competitive performance even with limited model capacity for enhanced parameter efficiency. 
Remembering Transformer learns a set of adapters to efficiently tackle various tasks, reducing the memory footprint through the adapter fusion. However, addressing longer task sequences necessitates learning hierarchical representations within these adapters. This would facilitate more flexible and efficient knowledge reuse for continual learning.

\section*{Acknowledgements}
We would like to thank Steve Lin and Zhirong Wu for their insightful discussions and valuable contributions to this work.

\bibliography{neurips24}
\bibliographystyle{plain}

\end{document}